\documentclass[letterpaper]{article} 
\usepackage{aaai25}  
\usepackage{times}  
\usepackage{helvet}  
\usepackage{courier}  
\usepackage[hyphens]{url}  
\usepackage{graphicx} 
\usepackage{natbib}  
\usepackage{caption} 
\usepackage{algorithm}
\usepackage{algorithmic}
\usepackage{graphicx}
\usepackage{booktabs}
\usepackage{tabularx}
\usepackage{diagbox} 
\usepackage{caption} 
\usepackage{siunitx}  
\usepackage{amsmath}
\usepackage{cleveref}
\usepackage{xspace}
\usepackage{arydshln}
\newcommand{\ie}{{\emph{i.e.}},\xspace}

\newcommand{\eg}{{\emph{e.g.}},\xspace}

\usepackage{newfloat}
\usepackage{listings}
\DeclareCaptionStyle{ruled}{labelfont=normalfont,labelsep=colon,strut=off} 
\floatstyle{ruled}
\newfloat{listing}{tb}{lst}{}
\floatname{listing}{Listing}
\title{SlerpFace: Face Template Protection via Spherical Linear Interpolation}

\author{
Zhizhou Zhong\textsuperscript{\rm 1}$\footnotemark[1]$, Yuxi Mi\textsuperscript{\rm 1}\thanks{Authors contributed equally to this paper.}, 
Yuge Huang\textsuperscript{\rm 2}\thanks{Corresponding authors.}, 
Jianqing Xu\textsuperscript{\rm 2}, 
Guodong Mu\textsuperscript{\rm 2}, 
Shouhong Ding\textsuperscript{\rm 2}, 
Jingyun Zhang\textsuperscript{\rm 3}, 
Rizen Guo\textsuperscript{\rm 3}, 
Yunsheng Wu\textsuperscript{\rm 2}, 
Shuigeng Zhou\textsuperscript{\rm 1}$\footnotemark[2]$
}
\affiliations{
\textsuperscript{\rm 1} Fudan University \\
\textsuperscript{\rm 2} Youtu Lab, Tencent \\
\textsuperscript{\rm 3} WeChat Pay Lab33, Tencent
\\

zzzhong22@m.fudan.edu.cn, \{yxmi20, sgzhou\}@fudan.edu.cn \\
 \{yugehuang, joejqxu, gordonmu, ericshding, simonwu\}@tencent.com \\
 \{naskyzhang, rizenguo\}@tencent.com
 }

\begin{document}

\maketitle

\begin{abstract}
Contemporary face recognition systems use feature templates extracted from face images to identify persons. To enhance privacy, face template protection techniques are widely employed to conceal sensitive identity and appearance information stored in the template. This paper identifies an emerging privacy attack form utilizing diffusion models that could nullify prior protection. 
The attack can synthesize high-quality, identity-preserving face images from templates, revealing persons' appearance. Based on studies of the diffusion model's generative capability, this paper proposes a defense by rotating templates to a noise-like distribution. 
This is achieved efficiently by spherically and linearly interpolating templates on their located hypersphere. 
This paper further proposes to group-wisely divide and drop out templates' feature dimensions, to enhance the irreversibility of rotated templates. 
The proposed techniques are concretized as a novel face template protection technique, SlerpFace. Extensive experiments show that SlerpFace provides satisfactory recognition accuracy and comprehensive protection against inversion and other attack forms, superior to prior arts.
\end{abstract}

%

\section{Introduction}
\label{sec:intro}
Face recognition (FR) is a biometric way to identify persons by facial appearance. Contemporarily, face recognition is enabled by comparing identity-discriminative feature vectors, or \textit{face templates}, extracted from face images via deep neural networks (DNN). 

Face templates are commonly considered sensitive data, as they carry identity and appearance information inferable of a specific person. To meet growing regulatory demands, face template protection (FTP) methods are proposed to conceal original templates, and securely represent them with an irreversible and revocable reference form~\cite{ISOIEC24745:2022}, known as \textit{protective templates}. These methods can be broadly divided into three categories: Crypto-based methods~\cite{boddeti2018secure,jindal2020secure,engelsma2022hers} process templates with encryption or security protocols in high latency and computation costs. Hash-based methods~\cite{mohan2019significant,dang2020fehash,kim2021ironmask,rathgeb2022deep,dusmanu2021privacy} turn templates into randomized codewords. Yet, they are less tolerant of minor facial attribute variations and hence could downgrade recognition. Recently, transform-based methods~\cite{phillips2019enhancing,abdellatef2020cancelable,hahn2022towards,shahreza2023mlp} have gained increasing attention as they are appealing in both accuracy and cost. They obtain protective templates via carefully designed transformations that obfuscate their binding with the original ones.

\begin{figure}[tpb]
    \centering
    \includegraphics[width=\linewidth]{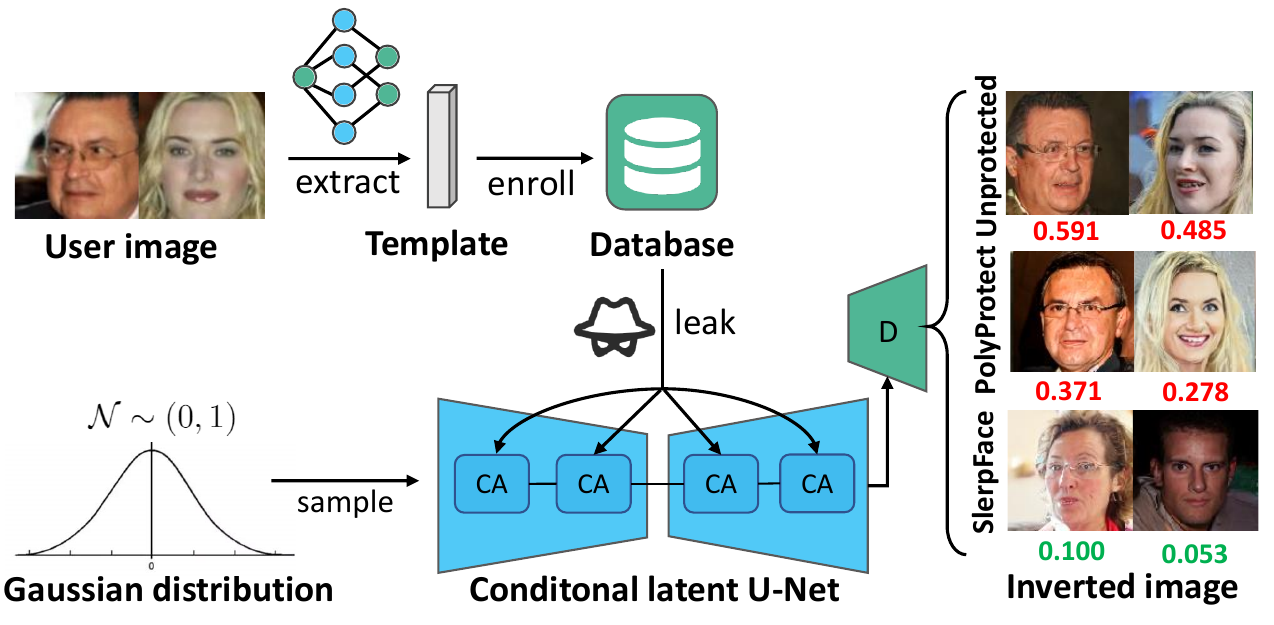}
    \caption{Paradigm of DM inversion attacks. 
    It receives templates as conditional contexts and synthesizes identity-descriptive images. 
    While unprotected templates and prior FTP arts are experimentally found vulnerable to inversion attacks, this paper presents a novel SlerpFace method as an effective defense. 
    It deteriorates DM to let it generate images with obfuscated facial semantics and lower similarity scores, hence preserving privacy.
    }
    \label{fig:inversion-attack-paradigm}
\end{figure}

Transform-based methods, however, could bear two privacy bottlenecks. First, they usually must pre-negotiate some secure parameters or \textit{secrets} for the transformation, which once exposed would compromise privacy. Second, they could also be susceptible to privacy attacks, where reconstruction~\cite{mai2018reconstruction,shahreza2022face,shahreza2023comprehensive} models using generative adversarial networks (GAN) or autoencoders (AE) and score-based techniques~\cite{razzhigaev2021darker,dong2023reconstruct,lai2021efficient,wang2021interpretable} may manage to recover partial facial appearance from protective templates, rendering protection less effective.

This paper further investigates an emerging type of attack, based on recent grave advances in diffusion models (DM)~\cite{ho2020denoising}. DM synthesizes high-quality images from randomly sampled noise through a learned denoising process, where the image’s content can be designated by an optional \textit{context} condition~\cite{lu2024coarse, ren2024hyper}. This paper identifies a new privacy threat from DM’s capability, referred to as \textit{inversion attacks}: Taking templates as the context, DM may invert face images that preserve the templates’ identity, hence revealing the persons’ face, as depicted in Fig.~\ref{fig:inversion-attack-paradigm}. Such DMs have been made possible by recent image synthesis arts~\cite{kansy2023controllable, Boutros2023IDiffFace}. This paper identifies inversion attacks as \textit{more dreadful than previous attack forms}, warranting special attention in future research: For the first time, it enables recovering both \textit{high-quality} and \textit{identity-preserving} face images from templates, imposing exacerbated threats. Prior transform-based arts are also proven vulnerable to inversion attacks, later demonstrated in Sec.~\ref{subsec-exp-inversion}.

To address privacy issues, this paper proposes a novel transform-based FTP method, SlerpFace. It effectively improves prior arts’ inadequate protection against secret exposure and different attacks, thus improving privacy.

Considering inversion attacks as the primary threat, this paper finds that DM's performance can deteriorate by altering the context's distribution. Specifically, when replacing authentic templates with randomly sampled Gaussian noise as context, DM is obfuscated from producing face images with a consistent identity, which is desirable for privacy. Drawing insights, this paper proposes to transform the original templates toward being alike sample-wise noises while maintaining discriminative identities. This is efficiently achieved by \underline{s}pherically and \underline{l}inearly int\underline{erp}olating (slerp) templates on their located hypersphere. The noises serve as secrets for transformation. This paper further addresses the noises' exposure by grouping and randomly dropping out protective templates' feature dimensions, where the division of groups is learnable to optimize recognizability. To the authors' knowledge, SlerpFace is the first FTP method to study resistance to inversion attacks. Extensive experiments suggest that it effectively prevents inversion and other attack forms as the protective templates are securely obfuscated. It also outperforms prior arts in accuracy and cost.

This paper's contributions are three-fold:
(1) It identifies the inversion attack as a severe privacy threat to transform-based template protection and analyzes the attack model's generative capability. 
(2) It suggests spherical linear interpolation as an effective defense, by rotating templates towards sample-wise noises. It further proposes feature grouping and dropout to enhance templates' privacy under secret exposure, and learnable feature grouping to improve recognizability. 
(3) It presents a novel FTP method, SlerpFace. Experiments demonstrate its superior privacy protection, better accuracy, and lower cost than prior arts.

\section{Related Work}\label{sec:related-work}
\subsection{Face Recognition}\label{subsec-rw-fr}
Modern FR systems recognize persons by comparing their face templates. Templates are feature vectors extracted from face images using DNNs, where angular margin losses~\cite{deng2019arcface, huang2020curricularface, kim2022adaface} are most employed during training to earn templates with identity discrepancy that facilitates recognition. They produce normalized templates that can be considered as unit vectors onto a hypersphere. During inference, cosine similarities are calculated to find the closest match.

\subsection{Face Template Protection}
This paper divides FTP methods into three branches: Crypto-based methods~\cite{boddeti2018secure,jindal2020secure,engelsma2022hers} use homomorphic encryption to turn templates into ciphertexts and perform calculations thereon. Their shortages involve high computation costs and reliance on the secrecy of encryption keys.

Hash-based methods use one-way schemes such as fuzzy commitment ~\cite{mohan2019significant,gilkalaye2019euclidean}, fuzzy vault~\cite{dong2021secure,rathgeb2022deep}, locality-sensitive hashing~\cite{dang2020fehash}, discretization~\cite{xu2020random}, and trainable models~\cite{chen2019deep} to map templates into irreversible protective codewords or hash values. Unfortunately, they often fail to achieve satisfactory recognition accuracy, as their means are less tolerant of the intra-identity variability inherent in facial attributes, resulting in false negatives. Recently, IronMask~\cite{kim2021ironmask} and ASE~\cite{dusmanu2021privacy} achieve protection with improved accuracy, by rotating templates to randomly chosen codewords and affining them into random subspace, respectively.

Transform-based methods apply task-specific transformations like feature reduction~\cite{pillai2011secure,hahn2022towards}, mixing~\cite{phillips2019enhancing,abdellatef2020cancelable} and rotation~\cite{shahreza2023mlp} to convert templates into protective forms, obscuring their bindings with the original ones. 
They differ from hash-based methods in must keep confidential some pre-negotiated secrets that designate the transformation. They could be susceptible to privacy attacks, as later experiments reveal. 
This paper proposes a novel transform-based method, SlerpFace, that can address the above inadequacies.

A research direction parallels to us is image protection~\cite{mi2023privacy, mi2024privacy, mi2022duetface, zhang2024pixelfade, zhang2024privacy,liu2024localfeaturesmeetstochastic,yuan2022pro,yuan2024pro}. They focus on the protection of the image transmitted to service providers, whereas SlerpFace is dedicated to ensuring the security of the stored face feature.

\subsection{Privacy Attacks}\label{subsec-rw-attack}
This paper studies attacks that attempt to recover facial appearances from templates and compromise privacy. 
They can be divided into three folds by attack means. 

Reconstruction attacks query the FR system with attacker-owned face images to obtain corresponding templates. 
They then use generative adversarial networks~\cite{truong2022vec2face,shahreza2023comprehensive, shahreza2023face} or autoencoders~\cite{cole2017synthesizing, mai2018reconstruction} trained on image-template pairs to learn an inverse fit that generates synthetic images. 

Score-based attacks~\cite{razzhigaev2021darker,vendrow2021realistic,dong2023reconstruct,lai2021efficient,wang2021interpretable,shahreza2023face} instigate through the knowledge of similarity scores. It recursively optimizes model-generated images by querying and maximizing their FR similarity scores with images in the database.

This paper further introduces inversion attacks, where diffusion models~\cite{Boutros2023IDiffFace, kansy2023controllable} receive templates as their contexts to generate images preserving the same identity, thus revealing persons' faces. Inversion attacks impose more dreadful threats as they can generate high-quality, identity-preserving images.

\section{Methodology}
\label{sec:method}

\begin{figure}[tpb]
    \centering
    \includegraphics[width=\linewidth]{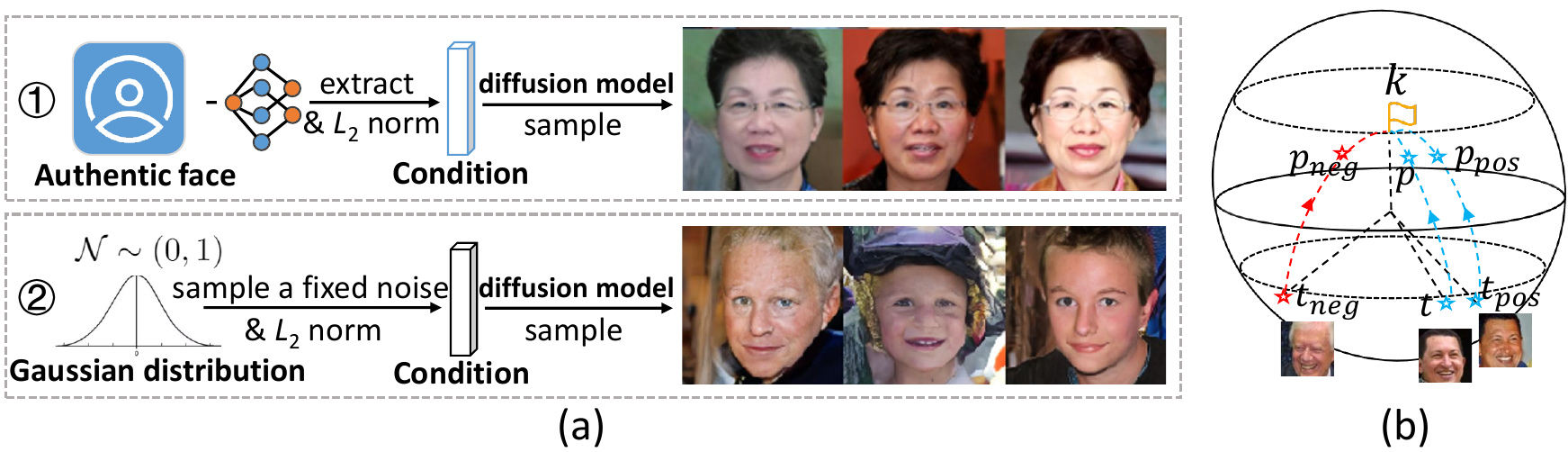}
    \caption{
    (a) DM's generative capability: Consistent identity faces from authentic templates, but deteriorates with noise templates, causing semantic variation. (b) Slerp rotation at $d$=3: Query, positive, and negative templates rotate in the same direction, maintaining margin differences.
    }
    \label{fig:dm-and-slerp}
\end{figure}

\subsection{Motivation}
\label{sec:motivation}

In practical applications, FR typically occurs between a server and its clients, with the server acting as an FR service provider that pre-trains the recognition model and enrolls identity templates to create a database. The database and model are shared with local devices or clients, who use them to recognize query faces locally. To protect templates' sensitive information, FTP's goal is to design a transformation
that turns database templates $\vec{t}$ into protective forms $\vec{p}$ via secret parameters $\vec{k}$, making them safer to share.

We consider inversion attacks as the primary threat to shared templates. A DM is a generative model concretized as $g: (\vec{\epsilon}, \vec{t})\rightarrow X$. Taking random Gaussian noise $\vec{\epsilon}$ as input and template $\vec{t}$ as context, it synthesizes a face image $X$ descriptive of $\vec{t}$'s identity, thus nullifying privacy. Testifying the attack's generative capability, we first train a DM using 
a pipeline from IDiff-Face~\cite{Boutros2023IDiffFace}. Then, we infer it multiple times with a \textit{fixed} template extracted from an authentic face image via a pre-trained FR model. The details of DM and its training are aligned with IDiff-Face. As shown in Fig.~\ref{fig:dm-and-slerp}(a1), the inverted images exhibit consistent facial appearances, \ie, the same elderly woman wearing glasses. This suggests that DM can generate identity-preserving face images of high quality.

Prior image synthesis arts~\cite{lugmayr2022repaint, Boutros2023IDiffFace} also use randomly sampled contexts as a practice to let DM generate unseen concepts, \eg new identities. In Fig.~\ref{fig:dm-and-slerp}(a2), we further choose a \textit{fixed} noise template that each of its feature dimensions is randomly sampled from Gaussian distribution. Taking it as DM's context, we find the inversion deteriorates: Though high-quality face images are still being generated, they no longer preserve a ``hypothesized'' identity but vary in semantics such as gender and age. We attribute the downgrade to the distributional discrepancy between noise and authentic templates. Studies~\cite{li2021spherical,shen2020interfacegan,chen2022geometry} prove that templates follow \textit{a priori} distributions that help preserve semantics. Randomly drawn noise most likely falls in a different distribution that is close to semantics' decision bounds, rendering uncertainties in images' facial appearance.

We leverage the observation as a means to prevent inversion attacks. Given that randomly sampled noise templates deteriorate the attack, an intuitive idea to protect templates is to ``move'' them toward the noise distribution, hence diminishing their discriminative semantics. 

\begin{figure*}[tpb]
    \centering
    \includegraphics[width=0.87\linewidth]{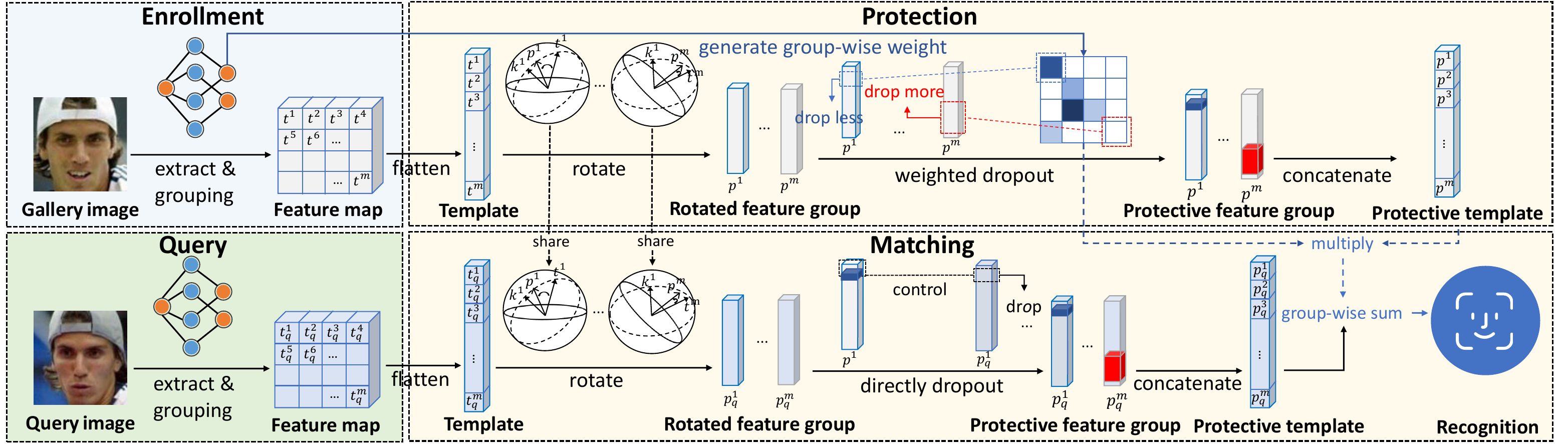}
    \caption{
    Pipeline of SlerpFace: (1) Train an FR model to extract and group feature maps as templates. (2) Protect templates by independently rotating feature groups toward a key template and applying random dropout based on learnable weights. (3) During inference, extract a query template. (4) Match it with enrolled templates using the same rotation and dropout.
    }
    \label{fig:pipeline}
\end{figure*}

\subsection{Rotation via Spherical Linear Interpolation}
\label{subsec-method-slerp}

Let $d$ be the feature dimension of templates. Recall in Sec.~\ref{subsec-rw-fr} that templates can be considered as unit vectors located on a $d$-dimensional hypersphere for most modern FR. To move template $\vec{t}$ toward noise distribution thus equals rotating $\vec{t}$ on the hypersphere to the direction of a noise template $\vec{k}$. We refer to $\vec{k}$ as \textit{key template} as it represents a secret that designates the rotation. Prior methods implement rotation of $\vec{t}$ by multiplying it with an orthogonal matrix $\vec{M}^{d\times d}$. However, to derive a random $\vec{M}$ is rather time-consuming, especially for large $d$~\cite{schreiber1989storage,chen2023devil}. Instead, we adopt an efficient way in light of a computer graphic study~\cite{shoemake1985animating}. It suggests that a 3D object can be smoothly rotated by interpolating its coordinates on a sphere. We refer to the technique as spherical linear interpolation, or \textit{slerp}. We generalize slerp to $d$-dimension and rotate $\vec{t}$ as:

\begin{equation}
    \vec{p} = \frac{\sin((1-\alpha)\theta)}{\sin\theta} \vec{t} + \frac{\sin(\alpha\theta)}{\sin\theta} \vec{k},
    \label{eq:slerp}
\end{equation}

\noindent where $\vec{p}$ denotes the rotated protective template and $\vec{k}$ is \textit{sample-wisely} chosen for each $\vec{t}$. $\theta = \arccos{\left( \frac{\vec{t}^\top \vec{k}}{\| \vec{t} \| \| \vec{k} \|}\right)}$ denotes the included angle between $\vec{t}$ and $\vec{k}$, and $\alpha$ is a hyper-parameter that controls the degree of rotation.

Based on previous discussions, we expect rotating $\vec{t}$ toward $\vec{k}$ to deteriorate inversion attacks by obfuscating DMs from generating face images aligned with $\vec{t}$'s identity. Section~\ref{subsec-exp-inversion} later testified to its effectiveness of protection.

Rotation also maintains templates' recognizability. Figure~\ref{fig:dm-and-slerp}(b) exemplifies the effect of slerp at $d$=3. Let $\vec{t},\vec{t}_{pos},\vec{t}_{neg}$ denote a query template and two templates with the same or different identities in the database, respectively. Initially, the angular margin between $\vec{t},\vec{t}_{pos}$ is smaller than that between $\vec{t},\vec{t}_{neg}$ as FR encourages templates to have small intra-class and large inter-class margins. Slerp rotates them toward key $\vec{k}$ by the same degree to obtain corresponding protective $\vec{p},\vec{p}_{pos},\vec{p}_{neg}$. We highlight that the relative difference among their margins is maintained as $\vec{p}_{pos}$ remains a closer match of $\vec{p}$. Hence, $\vec{p}$ will not be falsely recognized.
To apply slerp in practice, denote $\vec{P}=\{\vec{p}_1,\dots,\vec{p}_n\}$ as a set of $n$ database templates, protected by the server using corresponding keys $\vec{K}=\{\vec{k}_1,\dots,\vec{k}_n\}$. The server shares $\{\vec{P},\vec{K}\}$ with its clients. Given a query template $\vec{t}_q$, a client try rotating it with each $\vec{k}_{i\in[n]}$ to obtain $\vec{p}_{q_i}$ and match with $\vec{p}_i$. The relative marginal difference before and after rotation is maintained, hence the client can compare similarities to find the closest match.

\begin{figure*}[tpb]
    \centering
    \includegraphics[width=0.8\linewidth]{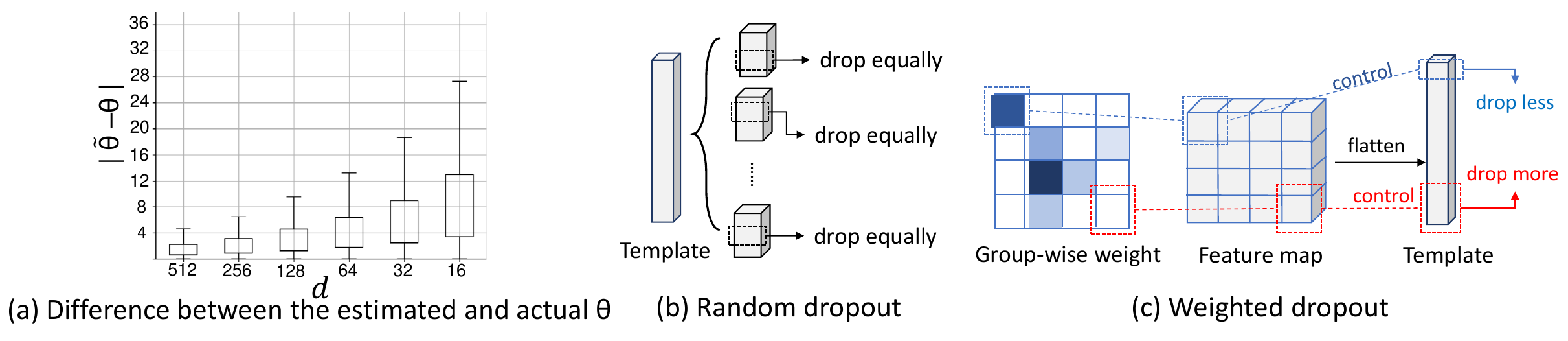}
    \caption{ (a) Ranges of $\Delta_{\theta}$ for templates with different feature dimensions $d$. $\Delta_{\theta}$ gradually increases as $d$ decreases. (b) Random dropout. Each feature group randomly discards equal dimensions of features. (c) Weighted dropout. Feature groups with larger weights discard fewer feature dimensions to better preserve crucial features.
    }\label{fig:delta-and-dropout}
\end{figure*}

\subsection{Feature Grouping and Dropout}
\label{subsec-method-grouping}

Rotating templates effectively gains resiliency against inversion attacks. Its protection is yet not intact, as we find the original $\vec{t}$ can still be recovered by any malicious client knowing of $\{\vec{p},\vec{k}\}$. Specifically, let $\{t_i,p_i,k_i\}_{i\in [d]}$ be the respective feature dimensions of $\{\vec{t},\vec{p},\vec{k}\}$. Equation~\ref{eq:slerp} can be rewritten as a full-rank linear system of $d$ equations with $d$ unknowns $\{t_1,\dots,t_d\}$:

\begin{equation}
    p_i = \frac{\sin((1-\alpha)\theta)}{\sin\theta} t_i + \frac{\sin(\alpha\theta)}{\sin\theta} k_i,\quad i \in [d].
    \label{eq:slerp-rewrite}
\end{equation}

\noindent The client can thus employ numerical calculation techniques, \eg the Newton-Raphson method~\cite{lindstrom1988newton}, to approximately solve Eq.~\ref{eq:slerp-rewrite} and obtain an estimated $\vec{\tilde{t}}\approx \vec{t}$. This will break the irreversibility of FTP and nullify its protection.

This section introduces a two-fold solution to enhance privacy. First, we intuitively observe that reducing the effective number of equations in Eq.~\ref{eq:slerp-rewrite} will make it an underdetermined system, leading to imprecise approximations of $\vec{\tilde{t}}$. We achieve the reduction by \textit{feature dropout}, \ie, randomly resetting a specific ratio $\beta$ of $\vec{p}$'s feature dimensions to 0. 

To evaluate the effectiveness of feature dropout, fixing a $\vec{k}$, $\vec{\tilde{t}}$'s precision can be quantified as its angle $\tilde{\theta}$ between $\vec{k}$, compared to $\theta$ between $\{\vec{t},\vec{k}\}$, \ie, $\Delta_{\theta}=|\tilde{\theta}-\theta|$. A larger $\Delta_{\theta}$ indicates $\vec{\tilde{t}}$ is further away from the authentic $\vec{t}$, which benefits privacy. Setting $\beta$=0.5, we experimentally draw $\vec{\tilde{t}}$ 10000 times for templates with different dimensions $d$, and depict their range of $\Delta_{\theta}$ in Fig.~\ref{fig:delta-and-dropout}(a). We find that feature dropout is less satisfactory for a large $d$ (\eg 512), as its $\Delta_{\theta} < 5^{\circ}$. On the other hand, $\Delta_{\theta}<28^{\circ}$ is salient when choosing a smaller $d$ (\eg 16), suggesting that the client's estimation of $\vec{\tilde{t}}$ is more uncertain.

Though a small $d$ would favor privacy, we cannot simply reduce $\vec{t}$'s dimension since such feature compression compromises recognizability heavily. Instead, we propose \textit{feature grouping} to leverage the findings, by dividing $\vec{t}$ into smaller groups and protecting each group separately. Specifically, let $\{\vec{t}^1,\dots,\vec{t}^m\}$ be $m$ equal-dimension groups divided from a $d$-dimension template $\vec{t}$ and $\{\vec{k}^1,\dots,\vec{k}^m\}$ be the division of its key $\vec{k}$. We normalize each group to a $(d/m)$-dimension hypersphere, and rotate it via slerp:

\begin{equation}
    \vec{p}^i = \frac{\sin((1-\alpha)\theta^i)}{\sin\theta^i} \vec{t}^i + \frac{\sin(\alpha\theta^i)}{\sin\theta^i} \vec{k}^i,\quad i\in [m],
    \label{eq:slerp-group}
\end{equation}

\noindent under angles $\vec{\theta}=\{\theta^i\}_{i\in [m]}$. We perform feature dropout on each of $\{\vec{p}^1,\dots,\vec{p}^m\}$, as shown in Fig.~\ref{fig:delta-and-dropout}(b), and concatenate them to form the protective $\vec{p}$. During inference, the similarity score is derived as the \textit{group-wise sum} similarity between a query $\vec{t}_q$ and database templates. 

\subsection{Learnable Feature Grouping}
\label{subsec-method-learnable}

\begin{figure*}[tpb]
    \centering
    \includegraphics[width=0.73\linewidth]{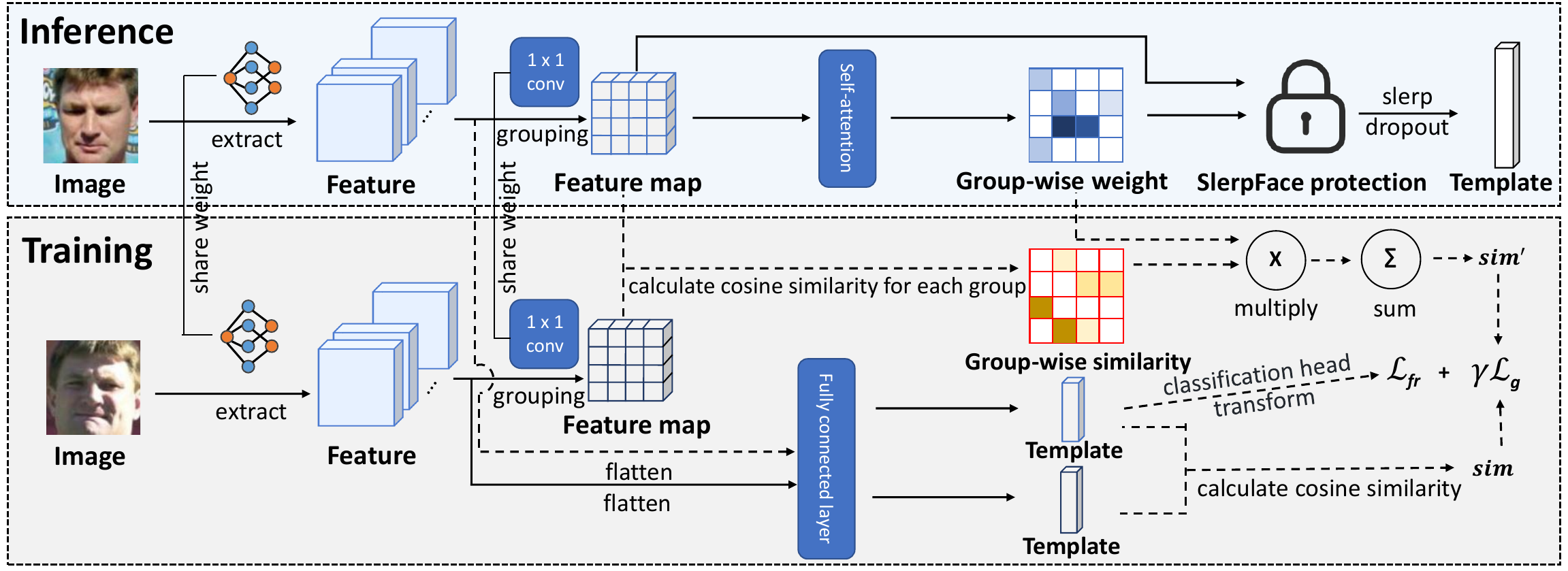}
    \caption{
    Paradigm of learnable feature grouping: Modify the FR model's output layer to generate feature maps, which are split into groups. Self-attention layers provide group-wise weights. For face image pairs, $\mathcal{L}_{g}$ aligns group-wise weighted similarities with original template similarities, trained alongside $\mathcal{L}_{fr}$. During inference, feature maps are reorganized as templates.}
    \label{fig:learnable-group}
\end{figure*}

We experimentally find feature grouping is at the cost of salient FR performance downgrade, later in Sec.~\ref{subsec-ablation}. To address this issue, we propose \textit{learnable feature grouping} inspired by a model interpretability study~\cite{lin2021xcos}. 
It suggests that facial recognizability can be regarded as the collective effort of different feature groups. 
Drawing insight, we aim to incorporate feature dimensions into learnable groups that achieve the same recognizability as the entire face together. 
Figure~\ref{fig:learnable-group} describes our approach.

Specifically, let $\{X_a, X_b\}$ be a pair of images with either the same or different identities. We train an FR model to extract their templates $\{\vec{t}_a,\vec{t}_b\}$ and denote their similarity as $\operatorname{sim}(\vec{t}_a,\vec{t}_b)$. To facilitate the model in producing appropriate groupings of features, we branch out before its final fully connection layer with a $1\times 1$ convolution layer to simultaneously obtain a pair of learnable feature maps $\{\vec{t}_{a'},\vec{t}_{b'}\}$, each with the shape of $(c,h,w)$. We regard each $c$-dimension spatial feature slice as a feature group, with a total of $m=hw$ groups. We calculate the similarity between each pair of feature groups $\{\vec{t}_{a'}^i,\vec{t}_{b'}^i\}$, denoted as $\operatorname{sim}(\vec{t}_{a'}^i,\vec{t}_{b'}^i)_{i\in[m]}$. We then calculate their weighted sum by group-wise weights $\vec{w}=\{w^1,\dots,w^m\}$, which are derived from the self-attention of feature maps. To encourage feature groups $\{\vec{t}_{a'}^i,\vec{t}_{b'}^i\}_{i\in[m]}$ together to be as recognizable as templates $\{\vec{t}_a,\vec{t}_b\}$, let $n$ be the batch size, we establish loss:
\begin{equation}
    \label{eq:group-loss}
    \mathcal{L}_{g} = \frac{1}{n} \sum_{i=1}^{n} \left\Vert \operatorname{sim}(\vec{t}_{a_i},\vec{t}_{b_i}) - \sum_{j=1}^{m} {w^j \operatorname{sim}(\vec{t}_{a'_i}^j,\vec{t}_{b'_i}^j)} \right\Vert_1,
\end{equation}
\noindent to bridge and align the weighted similarity sum to the template pair's similarity. The general FR loss (\eg ArcFace) is $\mathcal{L} = \mathcal{L}_{fr} + \gamma\mathcal{L}_{g}$, where $\gamma$ is a hyper-parameter. 

During inference, given a query image, we extract its learned feature map $\vec{t}'$, divide it into feature groups, and then concatenate all groups to form a flattened template, still denoted as $\vec{t}'$. 
$\vec{t}'$ is protected by group-wise rotation, as discussed in Sec.~\ref{subsec-method-grouping}. We further advocate an alternative \textit{weighted dropout}: As weights $\vec{w}$ reflect each group's contribution in calculating similarity, we propose to drop less or more feature dimensions for groups of $\vec{t}'$ with higher or lower weight, respectively, as shown in Fig.~\ref{fig:delta-and-dropout}(c). It helps preserve more crucial features compared to random dropout.

\section{Experiments}
\label{sec:experiments}

\subsection{Experimental Setup}
\label{subsec-exp-setup}

We employ an IR-50 model, trained on the MS1Mv2~\cite{guo2016ms} dataset on 8 GPUs in parallel with ArcFace loss as $\mathcal{L}_{fr}$, as the FR backbone. We train the model for 24 epochs using a stochastic gradient descent (SGD) optimizer, choosing the total batch size, initial learning rate, momentum, and weight decay as $256,0.01,0.9,0.0005$, respectively. We set parameters $(\alpha,\beta,\gamma,c,m)$ as $(0.9,0.5,1,16,49)$.
Evaluation is done on 5 regular-size datasets, LFW~\cite{lfwtechupdate}, CFP-FP~\cite{sengupta2016frontal}, AgeDB~\cite{moschoglou2017agedb}, CPLFW~\cite{zheng2018cross}, and CALFW~\cite{zheng1708cross}, and 2 large-scale datasets, IJB-B~\cite{whitelam2017iarpa} and IJB-C~\cite{maze2018iarpa}. 

\subsection{Recognition Performance}
\label{subsec-exp-accuracy}

\noindent \textbf{Compared methods.} 
We compare SlerpFace with an unprotected baseline and five FTP methods: \textbf{ArcFace} as the baseline; \textbf{Boddeti}~\cite{boddeti2018secure} using Fully Homomorphic Encryption; \textbf{IronMask}~\cite{kim2021ironmask} and \textbf{ASE}~\cite{dusmanu2021privacy}, both hash-based—IronMask employs orthogonal matrices for template rotation into random codewords, while ASE maps templates into random affine subspaces; and transform-based \textbf{MLP-Hash}~\cite{shahreza2023mlp} and \textbf{PolyProtect}~\cite{hahn2022towards}, with MLP-Hash rotating templates via pre-negotiated orthogonal matrices and PolyProtect translating templates into polynomials with specific exponents and coefficients.

\noindent \textbf{Recognizability and time cost.} We perform face recognition by verifying if two templates refer to the same person. We report \textit{accuracy} for LFW, CFP-FP, AgeDB, CPLFW, and CALFW, and \textit{TPR@FPR(1e-4)} for IJB-B and IJB-C. Results are summarized in Tab.~\ref{tab:performance}. Here, results for IronMask and MLP-Hash on IJB-B/C are marked as ``N/A'', as IronMask's approach to precisely match two codewords is viable only for accuracy results, and recognition for MLP-Hash on IJB-B/C takes prohibitive time.

Among the compared methods, FHE offers Bodddti the highest recognizability but is vulnerable if its key is compromised. Methods below the horizontal line can still provide protection when the key and template leakage. IronMask and ASE demonstrate subpar performance due to their hash-based methods' low tolerance for intra-identity facial attribute variations, impacting accuracy. SlerpFace surpasses previous models on most datasets, enhancing recognizability; however, it slightly trails PolyProtect on CFP-FP and CPLFW, which focus on facial pose variations. The accuracy gap is attributed to a reduction in descriptive capability from feature grouping, despite the groups being learnable. Nonetheless, note that the gap is marginal and serves as an efficient trade-off, as SlerpFace significantly surpasses PolyProtect in time cost and privacy.
 
The last two columns in Tab.~\ref{tab:performance} show the average enrollment (to register into a database) and matching (to match once with the database) time (ms) for a single template on a personal laptop, highlighting SlerpFace's advantage.

\subsection{Protection Against Privacy Attacks}
\label{subsec-exp-inversion}

Prior transform-based arts can be susceptible to privacy attacks, where attack models may reveal facial appearances from templates. We compare SlerpFace with transform-based MLP-Hash and PolyProtect using 3 SOTA attacks.

\noindent \textbf{Attack setup.} We set up one representative attack for each of the 3 attack forms discussed in Sec.~\ref{subsec-rw-attack}. (1) For inversion attacks, we designate IDiff-Face~\cite{Boutros2023IDiffFace}. It is a dataset synthesis work that uses latent DMs conditioned by templates to generate high-quality, identity-preserving images. We turn it into attacks using its DM's generative capability. (2) For score-based attacks, we consider EG3D~\cite{Chan2022}, which surpasses the classic score-based attack in recovery quality. It combines score-based techniques with reconstruction, optimizing the similarity scores of GAN-generated images. 
(3) For reconstruction attacks, we refer to NbNet~\cite{mai2018reconstruction}. It deconvolutes templates into face image form via a CNN structure. 

Using the LFW dataset as the source of ground truth images, we extract templates with a pre-trained FR model and secure them using MLP-Hash, PolyProtect, and SlerpFace. To conduct attacks, we train IDiff-Face, EG3D, and NbNet models from scratch for the unprotected baseline and three protection strategies, creating a total of 12 models. 

\begin{table*}[tpb]
    \small
    \centering
    \begin{tabular}{lcccccccccc} 
    \toprule
    \textbf{Method} & \textbf{LFW} & \textbf{CFP-FP} & \textbf{AgeDB} & \textbf{CALFW} & \textbf{CPLFW} & \textbf{IJB-B} & \textbf{IJB-C} & \textbf{Enrollment} & \textbf{Matching} \\
    \midrule
    ArcFace & 99.73 & 98.00 & 97.87 & 95.92 & 92.50 & 93.93 & 95.52 & - & - \\
    Boddeti & 99.73 & 97.86 & 97.81 & 95.84 & 92.41 & 93.88 & 95.47 & 1.99 & 22.4 \\
    \midrule
    IronMask & 84.42 & 52.70 & 53.22 & 50.00 & 50.00 & N/A & N/A & 832.22 & 0.25 \\
    ASE & 98.77 & 86.80 & 88.48 & 84.12 & 83.42 & 0.00 & 0.00 & 0.79 & 0.19 \\
    MLP-Hash & 98.82 & 91.40 & 93.67 & 93.07 & 87.12 & N/A & N/A & 134.31 & 131.76 \\
    PolyProtect & 99.30 & \textbf{94.00} & 95.28 & 94.77 & \textbf{89.22} & 87.37 & 89.96 & 1.86 & 1.88 \\
    \textbf{SlerpFace} & \textbf{99.42} & 92.79 & \textbf{95.70} & \textbf{94.82} & 88.90 & \textbf{89.96} & \textbf{92.25} & \textbf{0.35} & \textbf{0.17} \\
    \bottomrule
    \end{tabular}
    \caption{Comparison of recognition accuracy and time cost among SlerpFace, baseline, and SOTAs. 
    } 
    \label{tab:performance}
\end{table*}

 \begin{figure*}[tpb]
    \centering
    \includegraphics[width=\linewidth]{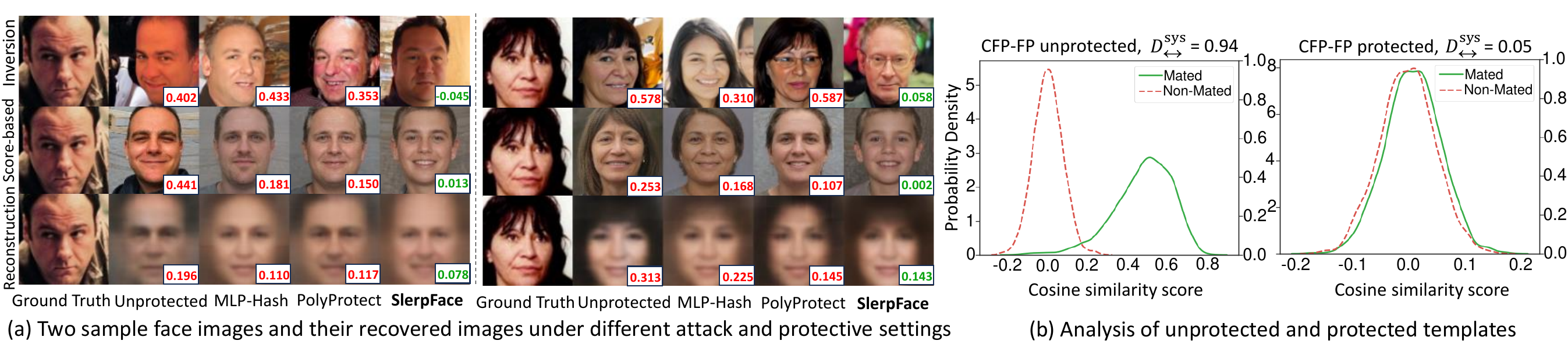}
    \caption{(a) Two sample face images and their recovered images under different attack and protective settings. 
    Values in the corner mark the similarity score.
    Dissimilar images and lower scores indicate better protection.
    (b) SSWL score between mated and non-mated template pairs for unprotected baseline and SlerpFace.
    } 
    \label{fig:privacy-attacks-unlink}
\end{figure*}

\begin{table}[t]
\centering
\small
\newcolumntype{Y}{>{\centering\arraybackslash}X}
\begin{tabularx}{\linewidth}{lYYYYYY}
\toprule
& \multicolumn{2}{c}{\textbf{Inversion}}
& \multicolumn{2}{c}{\textbf{Score-based}}
& \multicolumn{2}{c}{\textbf{Recon}}

\\
  & \multicolumn{1}{c}{Sim}    & \multicolumn{1}{c}{SRRA} & \multicolumn{1}{c}{Sim} & \multicolumn{1}{c}{SRRA} & \multicolumn{1}{c}{Sim} & \multicolumn{1}{c}{SRRA}  \\
  \midrule
    
Unprotected             & 0.46  & 98.83\% & 0.16 
& 17.15\%   &  0.12    &  6.50\%         \\

MLP-Hash & 0.19  & 27.55\% & 0.06    & 0.12\%    &  0.07     &  0.50\%
      \\
PolyProtect            & 0.32  & 73.83\% & 0.06      &  0.07\%   &0.07      & 0.52\%          \\

\textbf{SlerpFace}         & \textbf{0.05} & \textbf{0.10\%} & \textbf{0.05}   &  \textbf{0.06\%}    & \textbf{0.06}    & \textbf{0.43\%}   \\
\midrule
\end{tabularx}
\caption{Quantitative privacy analysis by Sim and SRRA. }
\label{tab:quant-measure}
\end{table}

\noindent \textbf{Attack evaluation.} Qualitatively, Fig.~\ref{fig:privacy-attacks-unlink}(a) exemplifies 2 sample ground truth images and the recovery from their unprotected and 3 method-corresponding protective templates. In corners, we mark the cosine similarity score between the templates of ground truth and recovered images. Dissimilar images and lower scores indicate better protection. Here, we highlight: (1) Inversion attacks impose more severe threats than other attack forms, as they generate both high-quality and identity-preserving images for unprotected baseline. (2) Prior arts fail to resist inversion as their recovered images display both high visual similarity and scores. SlerpFace provides better protection as its recoveries have low scores and show changes in semantics (\eg gender). (3) Though score-based and reconstruction attacks themselves exhibit lower capacities than inversion attacks, we can observe that SlerpFace outperforms prior arts with far lower scores. Notably, score-based attacks recover similar images for both samples protected by SlerpFace. We suspect this occurs as the protective templates do not provide effective semantics that influence the recovery, implying enhanced privacy.

Quantitatively, Tab.~\ref{tab:quant-measure} compares the dataset-wise similarity scores among different settings. We further quantify recovered images' quality by similarity (Sim) and the success rate of replay attacks (SRRA)~\cite{shahreza2023comprehensive}. It replays templates of recovered images as queries to determine if they can be recognized under a FAR threshold of 1e-3. Lower Sim and SRRA suggest better protection.

\subsection{Ablation Study}
\label{subsec-ablation}
\noindent \textbf{Component's contribution.} We proposed learnable feature grouping and weighted dropout to improve the recognizability of protective templates. In Tab.~\ref{tab:abla-accuracy}, ``w/o LW'' and ``w/o W'' denote SlerpFace without both learnable feature grouping and weighted dropout (equals Sec.~\ref{subsec-method-grouping} setting), and without weighted dropout alone, respectively. We observe that SlerpFace initially suffers salient accuracy downgrades due to feature grouping. Its accuracy is promoted for $5\%$ by learnable feature grouping and $2\%$ further by weighted dropout.

\subsection{Security Analysis}
\label{subsec-exp-security}

This section further discusses that SlerpFace satisfies important identity protection criteria~\cite{ISOIEC24745:2022}, known as \textit{irreversibility}, \textit{revocability}, and \textit{unlinkability}.

\begin{table}[t]
    \small
    \centering
    \begin{tabular*}{0.48\textwidth}{@{\extracolsep{\fill}}lccc} 
    \toprule
    \textbf{Method} & \textbf{LFW} & \textbf{CFP-FP} & \textbf{AGEDB} \\
    \midrule
    ArcFace & 99.73 & 98.00 & 97.87\\
    w/o LW & 99.37 & 87.13 & 90.73 \\
    w/o W & 99.00 & 92.56 & 93.67\\
    SlerpFace &\textbf{99.42} & \textbf{92.79} & \textbf{95.70} \\
    \bottomrule
    \end{tabular*}
    \caption{Components' role to SlerpFace's recognizability.} 
    \label{tab:abla-accuracy}  
\end{table}

\noindent \textbf{Irreversibility.} It requires that revealing the original templates $\vec{t}$ from the protective $\vec{p}$ should be computationally infeasible. 
A client with knowledge of $\vec{p}$ and key $\vec{k}$ might approximate $\vec{\tilde{t}} \approx \vec{t}$ using numerical methods like Newton-Raphson (NR)~\cite{lindstrom1988newton}, which approximately solve Eq.~\ref{eq:slerp-rewrite}.
We find that feature dropout hinders approximation by reducing the equations in Eq.~\ref{eq:slerp-rewrite}. 
Then, we design an NR-based attack to show SlerpFace's irreversibility and the effectiveness of feature dropout. We consider two protective settings with $(d,m)$=$(16,49)$. NR can be considered as a root-finding algorithm that produces successively better estimates $\vec{\tilde{t}}$ for $\vec{t}$. 
It begins with a random initial guess, succeeding with a close estimate or failing at maximum iterations, after which NR reinitializes and retries.

Taking the templates from LFW, CFP-FP, and AgeDB, we measure the cost by NR's average count of reruns $r$. To obtain results within finite time, we measure $r$ for each $d=16$ feature group. Hence, $r^m$ would be the ideal cost to recover full $\vec{\tilde{t}}$, as the attacker must succeed within each group \textit{simultaneously}. Without dropout, it takes NR $1.015^{49}\approx 2$ reruns to find a $\vec{\tilde{t}}$. However, with dropout, it takes around $3.6^{49}$ reruns, which is computationally infeasible. This suggests that dropout provides strong irreversibility.

Using the framework from Mai et al.~\shortcite{mai2020secureface}, we also \textit{theoretically} analyzed the irreversibility of SlerpFace, showing it has better entropy ($69.58$ compared to $59.41$) and matching accuracy ($86.20\%$ compared to $85.36\%$) on CFP-FP dataset. 

\noindent \textbf{Revocability.} It mandates that any compromised protective templates be revocable and replaceable. This can be easily achieved by re-enrolling $\vec{t}$ with a distinct key $\vec{k}'$.

\noindent \textbf{Unlinkability.} It requires that when generating different protective templates for the same person's identity, the generated templates cannot be associated with each other. We measure SlerpFace’s unlinkability via system score-wise linkability or SSWL score $D^{sys}_{\leftrightarrow}$, an evaluation metric proposed by Gomez-Barrero et al.~\shortcite{gomez2017general} and used in prior arts~\cite{mai2020secureface}. In essence, it is a $[0,1]$ value that measures the distributional discrepancy between template pairs describing the same or different identities, referred to as \textit{mated} or \textit{non-mated} pairs. To achieve unlinkability, the distributions of mated and non-mated pairs should be close and with small $D^{sys}_{\leftrightarrow}$. Figure~\ref{fig:privacy-attacks-unlink}(b) exhibits the pair-wise distribution for unprotected baseline and SlerpFace on the CFP-FP dataset. While baseline having $D^{sys}_{\leftrightarrow}$=0.94 indicates distinguishable distributions and undermines privacy, SlerpFace achieves $D^{sys}_{\leftrightarrow}$=0.05 and close distributions that are satisfactory for unlinkability.

\section{Conclusion}
\label{sec:conclusion}
This paper studies face template protection in face recognition. 
It first identifies diffusion-based inversion attacks as an exacerbated privacy threat.
 It then proposes a novel FTP method, SlerpFace, that effectively prevents inversion. 
SlerpFace rotates templates to a noise-like distribution that deteriorates attack models' capability, efficiently via spherical linear interpolation. 
It further proposes feature grouping and dropout, optimizable via a learnable approach, to enhance irreversibility. 
Extensive experiments demonstrate that SlerpFace outperforms SOTAs in both privacy protection and recognition performance.

\small
\bibliography{main}

\end{document}